%% file: main.tex
\documentclass[final]{cvpr}

\usepackage{times}
\usepackage{epsfig}
\usepackage{graphicx}
\usepackage{amsmath}
\usepackage{amssymb}

\usepackage[ampersand]{easylist}
\usepackage{comment}
\usepackage{algorithm}
\usepackage{algpseudocode}
\usepackage{bbm}
\usepackage{xcolor}
\usepackage{xargs}
\usepackage{tabularx}
\usepackage{fixmetodonotes}
\usepackage{caption}
\usepackage[title]{appendix}

\usepackage[symbol]{footmisc}

\usepackage{ifthen}
\newboolean{isCameraReady}
\setboolean{isCameraReady}{false}

\ifthenelse{\boolean{isCameraReady}}
{
  \usepackage{nopageno}
}{}

\usepackage[pagebackref=true,breaklinks=true,colorlinks,bookmarks=false]{hyperref}
\newtheorem{theorem}{Theorem}

\def\cvprPaperID{10610} % *** Enter the CVPR Paper ID here
\def\confYear{CVPR 2021}

\begin{document}

%%%%%%%%% TITLE
\title{SimPLE: Similar Pseudo Label Exploitation for Semi-Supervised Classification}

\author{Zijian Hu\thanks{Equal contributions; names ordered alphabetically.}
\quad Zhengyu Yang$^*$
\quad Xuefeng Hu
\quad Ram Nevatia\\
University of Southern California\\
{\tt\small \{zijianhu,yang765,xuefengh,nevatia\}@usc.edu}
% \and
}

\maketitle

%%%%%%%%% ABSTRACT
\begin{abstract}
\input{sections/abstract}
\end{abstract}

%%%%%%%%% BODY TEXT
\input{sections/introduction}

%-------------------------------------------------------------------------
\input{sections/related-work}

%-------------------------------------------------------------------------
\input{sections/method}

%-------------------------------------------------------------------------
\input{sections/experiments}

%-------------------------------------------------------------------------
\input{sections/conclusion}

%-------------------------------------------------------------------------

\section{Acknowledgement}
This material is based on research sponsored by Air Force Research Laboratory (AFRL) under agreement number FA8750-19-1-1000.
The U.S. Government is authorized to reproduce and distribute reprints for Government purposes notwithstanding any copyright notation therein.
The views and conclusions contained herein are those of the authors and should not be interpreted as necessarily representing the official policies or endorsements, either expressed or implied, of Air Force Laboratory, DARPA or the U.S. Government.

\clearpage
{\small
\bibliographystyle{ieee_fullname}
\bibliography{references}
}

\clearpage
\appendix
% \appendixpage
\addappheadtotoc
% \begin{appendices}
\input{sections/appendix}

% \end{appendices}

\end{document}

%% file: sections/abstract.tex
A common classification task situation is where one has a large amount of data available for training, but only a small portion is annotated with class labels.
The goal of semi-supervised training, in this context, is to improve classification accuracy by leverage information not only from labeled data but also from a large amount of unlabeled data.
%Semi-supervised learning leverages information not only from labeled data but also from a large amount of unlabeled data.
Recent works \cite{berthelot_mixmatch_2019,berthelot_remixmatch_2020, sohn_fixmatch_2020} have developed significant improvements by exploring the consistency constrain between differently augmented labeled and unlabeled data.
Following this path, we propose a novel unsupervised objective that focuses on the less studied relationship between the high confidence unlabeled data that are similar to each other.
The new proposed Pair Loss minimizes the statistical distance between high confidence pseudo labels with similarity above a certain threshold.
Combining the Pair Loss with the techniques developed by the MixMatch family \cite{berthelot_mixmatch_2019,berthelot_remixmatch_2020,sohn_fixmatch_2020}, our proposed SimPLE algorithm shows significant performance gains over previous algorithms on CIFAR-100 and Mini-ImageNet \cite{NIPS2016_MatchingNet}, and is on par with the state-of-the-art methods on CIFAR-10 and SVHN.
Furthermore, SimPLE also outperforms the state-of-the-art methods in the transfer learning setting, where models are initialized by the weights pre-trained on ImageNet\cite{krizhevsky_imagenet_2012} or DomainNet-Real\cite{Peng2019_DomainNet}.
The code is available at \href{https://github.com/zijian-hu/SimPLE}{github.com/zijian-hu/SimPLE}.

%% file: sections/introduction.tex
\section{Introduction}
\begin{figure}
    \centering
    \includegraphics[width=\columnwidth]{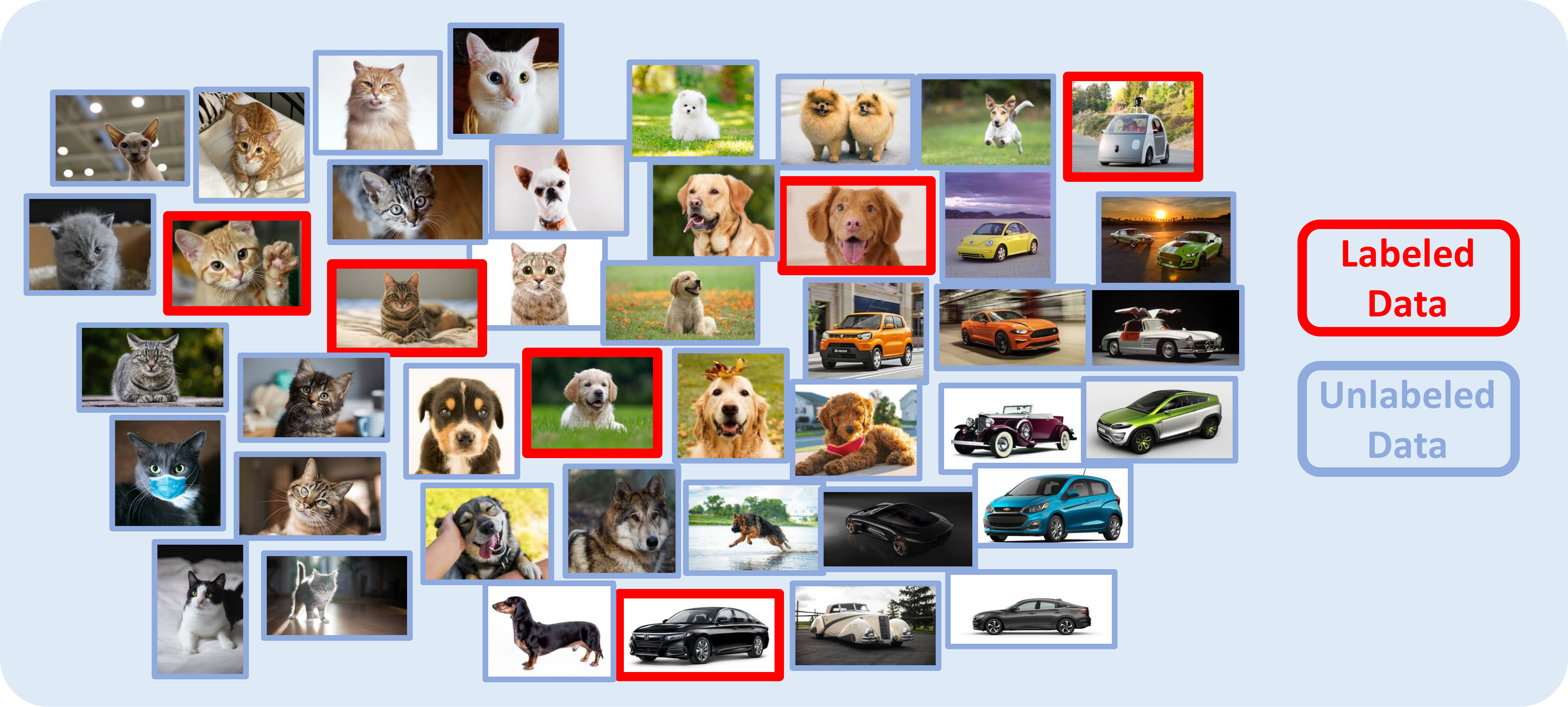}
    \caption{
        Illustration of an image set with a limited amount of labeled images among a large number of unlabeled images.
        Unlike unsupervised learning methods that only exploit the structure from unlabeled data,
        and supervised learning methods that only look at the limited amount of labeled data,
        semi-supervised learning utilizes information from both labeled and unlabeled data.
    }
    \label{fig:problem}
\end{figure}

Deep learning has recently achieved state-of-the-art performance on many computer vision tasks.
% The main strength of deep learning is that it generalizes well to large datasets.
One major factor in the success of deep learning is the large labeled datasets.
However, labeling large datasets is very expensive and often not feasible, especially in domains that require expertise to provide labels. Semi-Supervised Learning (SSL), on the other hand, can take advantage of partially labeled data, which is much more readily available, as shown in figure \ref{fig:problem}.

 %,as is one of the powerful approaches to alleviate the problem of limited amount of label data, by leveraging large amount of unlabeled data to improve the model performance.

A critical problem in semi-supervised learning is how to generalize the information learned from limited label data to unlabeled data. Following the \textit{continuity assumption}  that close data have a higher probability of sharing the same label \cite{chapelle_semi-supervised_2010}, many approaches have been developed \cite{szummer2002partially, zhou_learning_2003,douze2018low}, including the recently proposed Label Propagation \cite{Iscen2019_LabelPropagation}.

Another critical problem in semi-supervised learning is how to directly learn from the large amount of unlabeled data. Maintaining consistency between differently augmented unlabeled data has been recently studied and proved to be an effective way to learn from unlabeled data in both self-supervised learning \cite{chen2020simple, he_momentum_2020} and semi-supervised learning \cite{laine_temporal_2016, sajjadi_regularization_2016,berthelot_mixmatch_2019,berthelot_remixmatch_2020,sohn_fixmatch_2020,miyato_virtual_2019,tarvainen_mean_2017,zhang_wcp_2020,xie2020unsupervised}. Other than consistency regularization, a few other techniques have also been developed for the semi-supervised learning to leverage the unlabeled data, such as entropy minimization \cite{Miyato2019_VAT,lee2013pseudo,grandvalet2005semi} and generic regularization \cite{hinton1993keeping,loshchilov2017decoupled,zhang_mixup_2018,zhang2018three,verma_interpolation_2019}.

\begin{figure*}[ht]
    \centering
    \includegraphics[width=\textwidth]{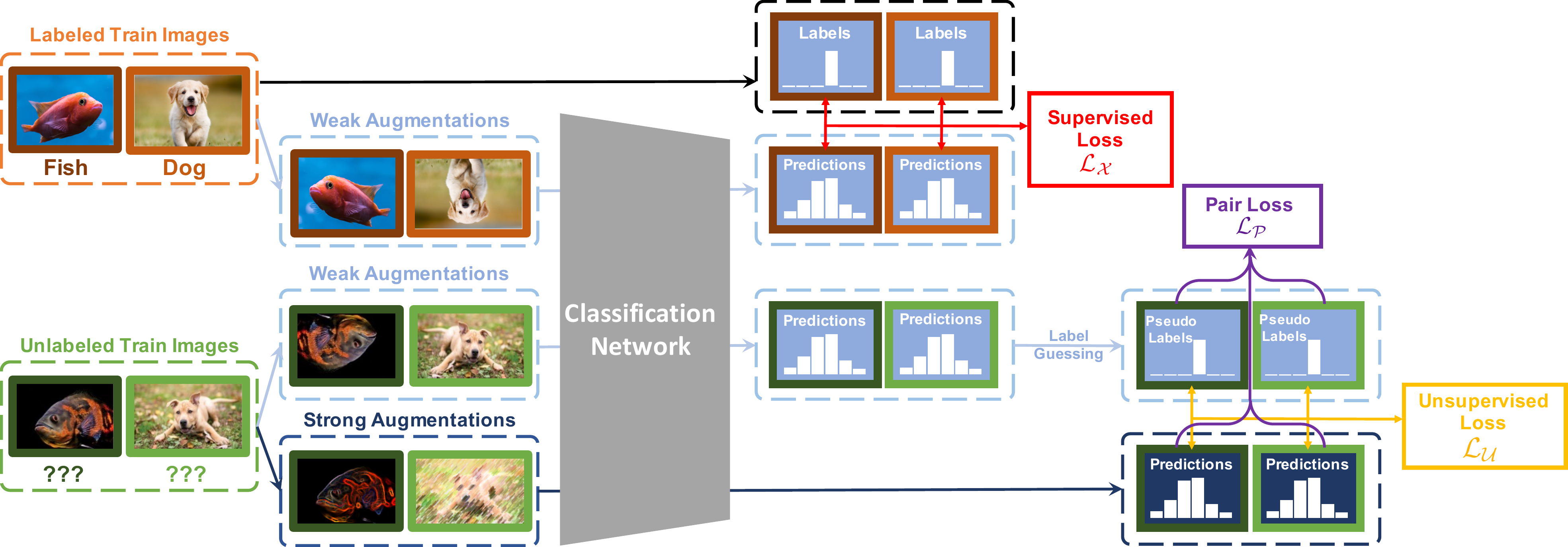}
    \caption{
        An overview of the proposed SimPLE algorithm. SimPLE optimizes the classification network with three training objectives:
        1) supervised loss $\mathcal{L_X}$ for augmented labeled data;
        2) unsupervised loss $\mathcal{L_U}$ that aligns the strongly augmented unlabeled data with pseudo labels generated from weakly augmented data;
        3) Pair Loss $\mathcal{L_P}$ that minimizes the statistical distance between predictions of strongly augmented data, based on the similarity and confidence of their pseudo labels.
    }
    \label{fig:overview}
\end{figure*}

The recently proposed MixMatch\cite{berthelot_mixmatch_2019} combined the above techniques and designed a unified loss function to let the model learn from differently augmented labeled and unlabeled data, together with the mix-up \cite{zhang_mixup_2018} technique, which encourages convex behavior between samples to increase models' generalization ability.
ReMixMatch \cite{berthelot_remixmatch_2020} further improves the MixMatch by introducing the Distribution Alignment and Augmentation Anchoring techniques, which allows the model to accommodate and leverage from the heavily augmented samples.
FixMatch \cite{sohn_fixmatch_2020} simplifies its previous works by introducing a confidence threshold into its unsupervised objective function and achieves state-of-the-art performance over the standard benchmarks.

However, while Label Propagation \cite{Iscen2019_LabelPropagation} mainly focuses on the relationship between labeled data to unlabeled data, and the MixMatch family \cite{berthelot_mixmatch_2019, berthelot_remixmatch_2020,sohn_fixmatch_2020} primarily focuses on the relationship between differently augmented unlabeled samples, the relationship between different unlabeled samples is less studied.

In this paper, we propose to take advantage of the relationship between different unlabeled samples.
% We introduce a novel Pair Loss, which encourages local clustering of predictions of high confidence.
% We introduce a novel Pair Loss, which encourages a pair of similar unlabeled samples to have similar predictions if at least one of them is of high confidence in its prediction.
% We introduce a novel Pair Loss, which.
We introduce a novel Pair Loss, which minimizes the distance between similar unlabeled samples of high confidence.

Combining the techniques developed by the MixMatch family \cite{berthelot_mixmatch_2019, berthelot_remixmatch_2020,sohn_fixmatch_2020}, we propose the SimPLE algorithm.
As shown in figure \ref{fig:overview}, the SimPLE algorithm generates pseudo labels of unlabeled samples by averaging and sharpening the predictions on multiple weakly augmented variations of the same sample.
Then, we use both the labels and pseudo labels to compute the supervised cross-entropy loss and unsupervised $L2$ distance loss.
These two terms push the decision boundaries to go through low-density areas and encourage consistency among different variations of the same samples.
Finally, with the newly proposed Pair Loss, we harness the relationships among the pseudo labels of different samples by encouraging consistency among different unlabeled samples which share a great similarity.

Our contribution can be described in four folds:
\begin{itemize}
    \item We propose a novel unsupervised loss term that leverages the information from high confidence similar unlabeled data pairs.
    \item Combining the techniques from MixMatch family \cite{berthelot_mixmatch_2019, berthelot_remixmatch_2020,sohn_fixmatch_2020} with the new Pair Loss, we developed the novel SimPLE algorithm for semi-supervised learning.
    \item We performed extensive experiments on the standard benchmarks and demonstrated the effectiveness of the proposed Pair Loss.
    SimPLE outperforms the state-of-the-art methods on CIFAR100 and Mini-ImageNet and on par with the state-of-the-art methods on CIFAR10, SVHN.
    \item We also evaluated our algorithm in a realistic setting where SSL methods are applied on pre-trained models, where the new proposed SimPLE algorithm also outperforms the current state-of-the-art methods.
\end{itemize}

%% file: sections/related-work.tex
\section{Related Work}
\label{section:related-work}

\subsection{Consistency Regularization}
Consistency regularization is widely used in the field of SSL. It refers to the idea that a model's response to an input should remain consistent, when  perturbations are used on the input or the model. The idea is first proposed in \cite{laine_temporal_2016, sajjadi_regularization_2016}. In its simplest form, the regularization can be achieved via the loss term:
\begin{equation}
    \| \mathrm{p}_\text{model}(y | A(x); \theta) - \mathrm{p}_{\text{model}^\prime}(y | A(x); \theta) \|_2^2
\end{equation}
 The stochastic transformation $A(x)$ can be either domain-speciﬁc data augmentation \cite{berthelot_mixmatch_2019, laine_temporal_2016, sajjadi_regularization_2016, berthelot_remixmatch_2020}, drop out \cite{sajjadi_regularization_2016}, random max pooling\cite{sajjadi_regularization_2016}, or adversarial transformation \cite{miyato_virtual_2019}. A further extension of this idea is to ``perturb'' the model,  $\mathrm{p}_{\text{model}^\prime}$, instead of the input. The perturbation can be a time ensembling of model at different time step \cite{laine_temporal_2016, tarvainen_mean_2017}, or an adversarial perturbation on model's parameter $\theta$ \cite{zhang_wcp_2020}. Also, many works choose to minimize cross entropy instead of the $L_2$ norm \cite{miyato_virtual_2019, xie2020unsupervised, berthelot_remixmatch_2020, sohn_fixmatch_2020}.

\subsubsection{Augmentation Anchoring}
 Augmentation Anchoring is first proposed by ReMixMatch \cite{berthelot_remixmatch_2020} and further developed in FixMatch \cite{sohn_fixmatch_2020}.
 It is a form of consistency regularization that involves applying different levels of perturbations to the input.
 A model's response to a slightly perturbed input is regarded as the ``anchor'', and we try to align model's response to a severely perturbed input to the anchor.
 For example, we can slightly perturb the input by applying an ``easy'' augmentation such as horizontal flipping and severely perturb the input by applying a ``hard'' augmentation such as Gaussian blurring.
 As the model's response to a slightly perturbed input is less unstable, including Augmentation Anchoring increases the stability of the regularization process.

\subsection{Pseudo-labeling}
% Use FixMatch
Pseudo labels are artificial labels generated by the model itself and are used to further train the model.
Lee \cite{lee2013pseudo} picks the class with the highest predicted probability by the model as the pseudo label.
However, pseudo labels are only used during the fine-tuning phase of the model, which has been pre-trained.
When we minimize the entropy on pseudo labels, we encouraged decision boundaries among clusters of unlabeled samples to be in the low-density region, which is requested by \textit{low-density separation assumption} \cite{chapelle_semi-supervised_2010}.
In this paper, for simplicity, we use a single lower case letter, $p \sim \Delta^N$ (the $N$-probability simplex), to represent either a hard label (a one-hot vector) or a soft label (a vector of probabilities).
A simple yet powerful extension of pseudo-labeling is to filter pseudo labels based on a confidence threshold \cite{french_self-ensembling_2018, sohn_fixmatch_2020}.
We define the confidence of a pseudo label as the highest probability of it being any class (i.e., $\max_{i}(p_i)$).
For simplicity, from now on, we will use $\max (p)$ as a shorthand notation for the confidence of any label $p$.
With a predefined confidence threshold $\tau_c$, we reject all pseudo labels whose confidence is below the threshold (i.e., $\max(p)<\tau_c$).
The confidence threshold allows us to focus on labels with high confidence (low entropy) that are away from the decision boundaries.

\subsection{Label Propagation}
Label propagation is a graph-based idea that tries to build a graph whose nodes are the labeled and unlabeled samples, and edges are weighted by the similarity between those samples \cite{chapelle_semi-supervised_2010}.
Although it is traditionally considered as a transductive method \cite{szummer2002partially, zhou_learning_2003}, recently, it has been used in an inductive setting as a way to give pseudo labels.
In \cite{Iscen2019_LabelPropagation}, the authors measure the similarity between feature representations of labeled and unlabeled samples embedded by a CNN.
Then, each sample is connected with the $K$ neighbors with the highest similarity to construct the affinity graph.
After pre-training the model in a supervised fashion, they train the model and propagate the graph alternatively.
The idea of using $K$ nearest neighbors to build a graph efficiently is proposed in \cite{douze2018low}, as most edges in the graph should have a weight close to $0$.
Our similarity threshold, $\tau_c$, takes a similar role.

%% file: sections/method.tex
% new commands
\newcommand{\funcHelper}[2]{
    \ensuremath{\operatorname{#1}\left(#2\right)}
}

\newcommandx{\modelPred}[3][1=\tilde{y}, 3=\theta]{
    \ensuremath{\mathrm{p}_{\text{model}}\left(#1 \mid #2 ; #3\right)}
}

\newcommandx{\modelEMAPred}[3][1=\tilde{y}, 3=\theta]{
    \ensuremath{\mathrm{p}_{\text{model}^\prime}\left(#1 \mid #2 ; #3\right)}
}

\newcommand{\simFunc}[2]{
    \ensuremath{f_{\operatorname{sim}}\left(#1, #2\right)}
}

\newcommand{\distFunc}[2]{
    \ensuremath{f_{\operatorname{dist}}\left(#1, #2\right)}
}

\newcommandx{\thresholdFunc}[2]{
    \ensuremath{\varphi_{#2}\left(#1\right)}
}

\newcommand{\crossEntropy}[2]{
    \ensuremath{H\left(#1, #2\right)}
}

\newcommandx{\norm}[3][2=, 3=F]{
    \ensuremath{\left\| #1 \right\|^{#2}_{#3}}
}

\newcommandx{\indicatorFunc}[3][2=>]{
    \ensuremath{\mathbbm{1}_{\left(#1 #2 #3\right)}}
}

\begin{algorithm*}
    \caption{SimPLE algorithm}
    \label{algorithm:SimPLE}
    \begin{algorithmic}[1]
        \State \textbf{Input}:
        Batch of labeled examples and their one-hot labels $\mathcal{X}=\left( \left(x_b, y_b\right);b\in\left( 1,\dots,B \right) \right)$,
        batch of unlabeled examples $\mathcal{U}=\left( u_b;b\in\left( 1,\dots,B \right) \right)$,
        sharpening temperature $T$,
        number of weak augmentations $K$,
        number of strong augmentations $K_{\text{strong}}$,
        confidence threshold $\tau_c$,
        similarity threshold $\tau_s$.

        \For{$b=1$ to $B$}
            \State $\tilde{x}_b = A_\text{weak}\left(x_b\right)$
            \Comment{Apply weak data augmentation to $x_b$}
            \For{$k=1$ to $K$}
                \State $\tilde{u}_{b,k} = A_\text{weak}\left(u_b\right)$
                \Comment{Apply $k^\mathrm{th}$ round of weak data augmentation to $u_b$}
            \EndFor

            \For{$k=1$ to $K_{\text{strong}}$}
                \State $\hat{u}_{b,k} = A_\text{strong}\left(u_b\right)$
                \Comment{Apply $k^\mathrm{th}$ round of strong augmentation to $u_b$}
            \EndFor

            \State $\bar{q}_b=\frac{1}{K} \sum^{K}_{k=1} \modelEMAPred{\tilde{u}_{b, k}}$
            \Comment{Compute average predictions across all \textbf{weakly} augmented $u_b$ using EMA}

            \State $q_b = \operatorname{Sharpen}(\bar{q}_b,T)$
            \Comment{Apply temperature sharpening to the average prediction}
        \EndFor

        \State $\hat{\mathcal{X}} = \left( \left(\tilde{x}_b, y_b\right);b\in\left( 1,\dots,B \right) \right)$
        \Comment{Weakly augmented labeled examples and their labels}

        \State $\hat{\mathcal{U}} = \left( \left(\hat{u}_{b, k}, q_b\right);b\in\left( 1,\dots,B \right), k\in\left( 1,\dots,K_{\text{strong}} \right) \right)$
        \Comment{Strongly augmented unlabeled examples, guessed labels}

        % compute losses
        \State $ \mathcal{L_X} = \frac{1}{\left| \mathcal{X}' \right|} \sum_{x,y \in \hat{\mathcal{X}}} \crossEntropy{y}{\modelPred{x}} $
        \Comment{Compute supervised loss}

        \State $ \mathcal{L_U} = \frac{1}{L \left| \hat{\mathcal{U}} \right|}
            \sum_{u,q \in \hat{\mathcal{U}}}
                \indicatorFunc{\max\left(q\right)}{\tau_c}
                \norm{q - \modelPred{u}}[2][2] $
        \Comment{Compute thresholded unsupervised loss}

        \State $ \mathcal{L_P} = \funcHelper{PairLoss}{
            \hat{\mathcal{U}},
            \tau_c,
            \tau_s
        } $
        \Comment{Compute Pair Loss}

        % return final loss
        \State \textbf{return} $ \mathcal{L_X} +
                    \lambda_{\mathcal{U}} \mathcal{L_U}
                    + \lambda_{\mathcal{P}} \mathcal{L_P} $
        \Comment{Compute loss $\mathcal{L}$ from $\hat{\mathcal{X}}$ and $\hat{\mathcal{U}}$}
    \end{algorithmic}
\end{algorithm*}

\section{Method}
To take full advantage of the vast quantity of unlabeled samples in SSL problems, we propose the SimPLE algorithm that focuses on the relationship between unlabeled samples.
In the following section, we first describe the semi-supervised image classification problem. Then, we develop the major components of our methods and incorporate everything into our proposed SimPLE algorithm.

\subsection{Problem Description}
We define the semi-supervised image classification problem as following. In a $L$-class classification setting, let $\mathcal{X}=\left(\left(x_{b}, y_{b}\right) ; b \in(1, \ldots, B)\right)$ be a batch of labeled data, and $\mathcal{U}=\left(u_{b} ; b \in(1, \ldots, B)\right)$ be a batch of unlabeled data. Let $\modelPred{x}$ denote the model's predicted softmax class probability of input $x$ parameterized by weight $\theta$.

\subsection{Augmentation Strategy}
Our algorithm uses Augmentation Anchoring \cite{berthelot_remixmatch_2020, sohn_fixmatch_2020}, in which pseudo labels come from weakly augmented samples act as ``anchor'', and we align the strongly augmented samples to the ``anchor''.
Our weak augmentation, follows that of MixMatch\cite{berthelot_mixmatch_2019}, ReMixMatch \cite{berthelot_remixmatch_2020}, and FixMatch \cite{sohn_fixmatch_2020}, contains a random cropping followed by a random horizontal flip.
We use RandAugment \cite{Cubuk2020_RandAugment} or a fixed augmentation strategy that contains difficult transformations such as random affine and color jitter as strong augmentation.
For every batch, RandAugment randomly selects a fixed number of augmentations from a predefined pool; the intensity of each transformation is determined by a magnitude parameter.
In our experiments, we find that method can adapt to high-intensity augmentation very quickly. Thus, we simply fix the magnitude to the highest value possible.

\subsection{Pseudo-labeling}
Our pseudo labeling is based on the label guessing technique used in \cite{berthelot_mixmatch_2019}. We first take the average of the model's predictions of several weakly augmented versions of the same unlabeled sample as its pseudo label.
As the prediction is averaged from $K$ slight perturbations of the same input instead of $K$ severe perturbation \cite{berthelot_mixmatch_2019} or a single perturbation \cite{berthelot_remixmatch_2020, sohn_fixmatch_2020}, the guessed pseudo label should be more stable.
Then, we use the sharpening operation defined in \cite{berthelot_mixmatch_2019} to increase the temperature of the label's distribution:
\begin{equation}
    \operatorname{Sharpen}(p, T):= \frac{p^{\frac{1}{T}}}{\textbf{1}^\top p^{\frac{1}{T}}}
\end{equation}
As the peak of the pseudo label's distribution is ``sharpened'', the network will push this sample further away from the decision boundary.
Additionally, following the practice of MixMatch \cite{berthelot_mixmatch_2019}, we use the exponential moving average of the model at each time step to guess the labels.

\subsection{Loss}
Our loss consists of three terms:  the supervised loss $\mathcal{L_X}$, the unsupervised loss $\mathcal{L_U}$, and the Pair Loss $\mathcal{L_P}$.
\begin{align}
    \mathcal{L} &= \mathcal{L_X} + \lambda_{\mathcal{U}} \mathcal{L_U} + \lambda_{\mathcal{P}} \mathcal{L_P} \\
    \mathcal{L_X} &= \frac{1}{\left| \mathcal{X}' \right|} \sum_{x,y \in \hat{\mathcal{X}}} \crossEntropy{y}{\modelPred{x}} \\
    \mathcal{L_U} &= \frac{
        \sum_{u,q \in \hat{\mathcal{U}}}
        \indicatorFunc{\max\left(q\right)}{\tau_c}
        \norm{q - \modelPred{u}}[2][2]
        }{L \left| \hat{\mathcal{U}} \right|}
\end{align}
$\mathcal{L_X}$ calculates the cross-entropy of weakly augmented labeled samples; $\mathcal{L_U}$ represents the $L_2$ distance between strongly augmented samples and their pseudo labels, filtered by confidence threshold.
Notice that $\mathcal{L_U}$ only enforces consistency among different perturbations of the same samples but not consistency among different samples.
\subsubsection{Pair Loss}
\begin{figure}
    \centering
    \includegraphics[width=\columnwidth]{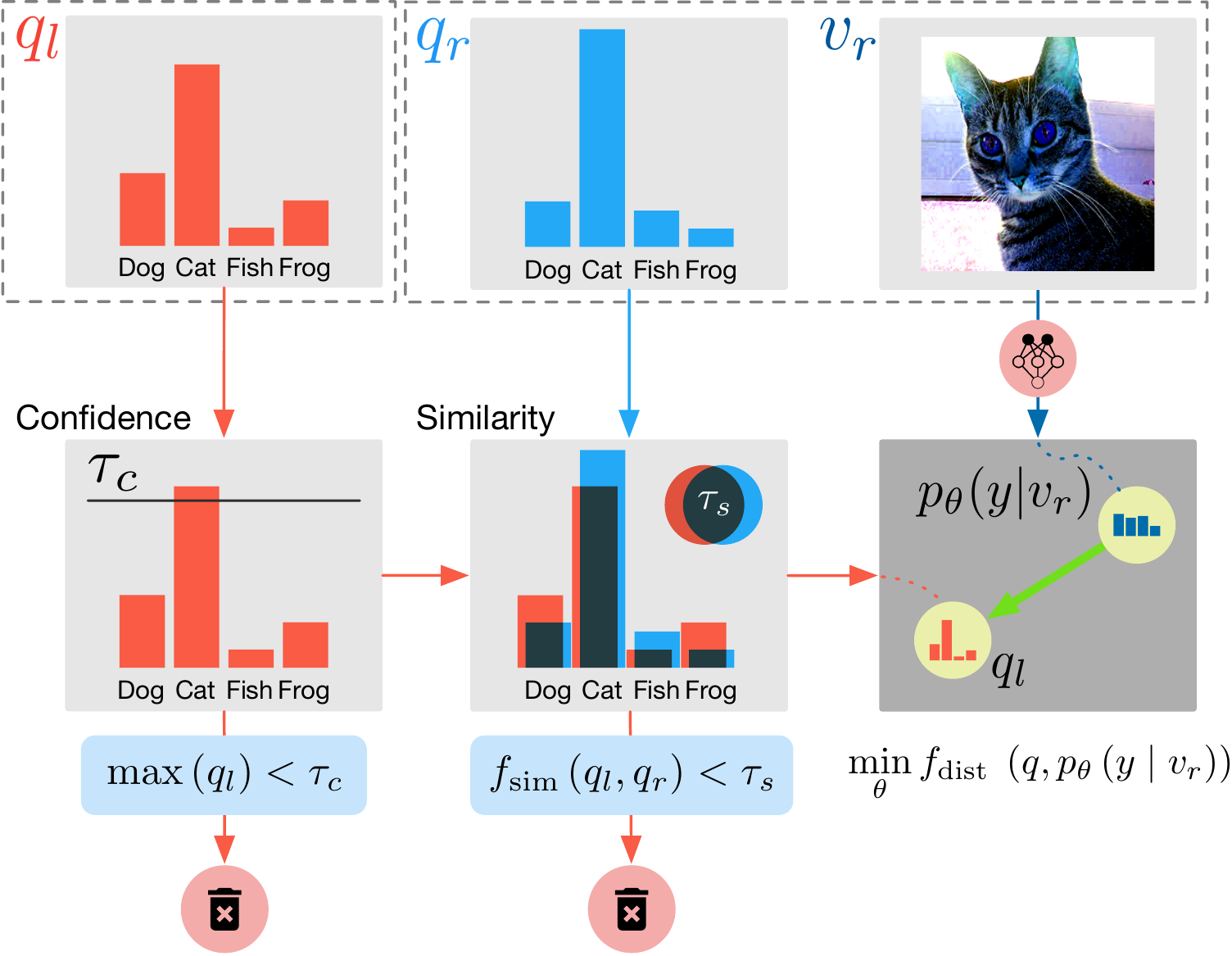}
    \caption{
        Pair Loss Overview.
        Given a pseudo label $q_l$ (red) which is a probability vector representing the guessed class distribution, if the highest entry in $q_l$ surpasses the confidence threshold $\tau_c$, $q_l$ will become an ``anchor''.
        Then, for any pseudo label and image tuple $q_r$ (light blue) and $v_r$ (dark blue), if the overlapping proportion (i.e. similarity) between $q_l$ and $q_r$ is greater than the confidence threshold $\tau_s$, this tuple $(q_r, v_r)$ will contribute toward the Pair Loss by pushing model's prediction of a strongly augmented version of $v_r$ to the ``anchor'' $q_l$ (green arrow).
        During this process, if either threshold can not be satisfied, $q_l, q_r, v_r$ will be rejected.
    }
    \label{fig:pair-loss}
\end{figure}

As we aim to exploit the relationship among unlabeled samples, we hereby introduce a novel loss term, Pair Loss, that allows information to propagate implicitly between different unlabeled samples.
In Pair Loss, we use a high confidence pseudo label of an unlabeled point, $p$, as an ``anchor.''
All unlabeled samples whose pseudo labels are similar enough to $p$ need to align their predictions under severe perturbation to the ``anchor.''
Figure \ref{fig:pair-loss} offers an overview of this selection process. During this process, the similarity threshold ``extended'' our confidence threshold in an adaptive manner, as a sample whose pseudo label confidence is below the threshold can still be selected by the loss and be pushed to a higher confidence level.
Formally, we defined the Pair Loss as following:
\begin{equation}
    \label{eqn:pairLoss}
    \begin{aligned}
        \mathcal{L_P} &= \frac{1}{\binom{K'B}{2}}
        \sum_{
            \substack{
                i,j \in \left[\left|\mathcal{U}'\right|\right], i \ne j \\
                \left(v_l, q_l\right) = \mathcal{U}'_{i}\\
                \left(v_r, q_r\right) = \mathcal{U}'_{j}
            }
        }
        \thresholdFunc{\max\left(q_l\right)}{\tau_c} \\
        &\cdot \thresholdFunc{\simFunc{q_l}{q_r}}{\tau_s} \\
        &\cdot \distFunc{q_l}{\modelPred{v_r}}
    \end{aligned}
\end{equation}
Here, $\tau_c$ and $\tau_s$ denote the confidence threshold and similarity threshold respectively. $\varphi_t(x)=\indicatorFunc{x}{t} x$ is a hard threshold function controlled by threshold $t$.
% \begin{align*}
%     \varphi_t(x) &=
%         \begin{cases}
%             x & \text{if } x > t \\
%             0 & \text{otherwise}
%         \end{cases}
% \end{align*}
$\simFunc{p}{q}$ measures the similarity between two probability vectors $p, q$ by Bhattacharyya coefficient \cite{Bhattacharyya1946_BC}.
The coefficient is bounded between $[0, 1]$, and represents the size of the overlapping portion of the two discrete distributions:
\begin{equation}
    \simFunc{p}{q} = \sqrt{p}^\top \sqrt{q}
\end{equation}
$\distFunc{p}{q}$ measures the distance between two probability vectors $p, q$.
As $\simFunc{p}{q} \in [0, 1]$, we choose the distance function to be $\distFunc{p}{q} = 1 - \simFunc{p}{q}$.

%\TODO{need it or not}
Although based on analysis, we found that $\cos(\cos^{-1}(\sqrt{\tau_c})+\cos^{-1}(\tau_s))^2$ is the infimal confidence a label need to have for it to be selected by both thresholds, such low confidence label are rarely selected in practice.
Based on empirical evidence, we believe this is caused by the fact a label $p$ that can pass through the high confidence threshold typically has a near one-hot distribution.
Thus, for another label $q$ to fall in the similarity threshold of $q$, it must also have relatively high confidence. Due to this property, the Pair Loss is not very sensitive to the choices of hyperparameters $\tau_s$, $\tau_c$, which we will show empirically in section \ref{section:ablation}.

\subsubsection{Motivation for Different Loss Formulations}
We follow MixMatch \cite{berthelot_mixmatch_2019} in choosing supervised loss $\mathcal{L}_{\mathcal{X}}$ and unsupervised loss $\mathcal{L}_{\mathcal{U}}$ terms. We use the Bhattacharyya coefficient \cite{Bhattacharyya1946_BC} in our Pair Loss because it measures the overlap between two distributions and allows a more intuitive selection of the similarity threshold $\tau_s$. Although we believe that the Bhattacharyya coefficient \cite{Bhattacharyya1946_BC} is more suitable than $L_2$ distance (or $2 - L_2$) to measure the similarity between two distributions, we keep the $L_2$ distance in unsupervised loss term to provide a better comparison with MixMatch  \cite{berthelot_mixmatch_2019}.
Moreover, as cross-entropy measures the entropy and is asymmetric, it is not a good distance measurement between distributions.
In our experiments, we observe that SimPLE with $L_2$ Pair Loss has $0.53\%$ lower test accuracy than the original.

\subsection{SimPLE Algorithm}
By putting together all the components introduced in this section, we now present the SimPLE algorithm.
During training, for a mini-batch of samples, SimPLE first augment labeled and unlabeled samples with both weak and strong augmentations.
The pseudo labels of the unlabeled samples are obtained by averaging and then sharpening the models' predictions on the weakly augmented unlabeled samples.
Finally, we optimize the loss terms based on augmented samples and pseudo labels.
During testing, SimPLE uses the exponential moving average of the weights of the model to make predictions, as the way done by MixMatch in \cite{berthelot_mixmatch_2019}.
Figure \ref{fig:overview} gives an overview of SimPLE, and the complete algorithm is in algorithm \ref{algorithm:SimPLE}.

%% file: sections/experiments.tex
\input{tables/cifar100}
\input{tables/cifar10-svhn}
\input{tables/mini-imagenet}

\section{Experiments}
\label{section:experiments}

% \subsection{Implementation Detail}

Unless specified otherwise, we use Wide ResNet 28-2 \cite{BMVC2016_WRN} as our backbone and AdamW \cite{loshchilov2018_AdamW} with weight decay for optimization in all experiments. We also use the exponential moving average (EMA) of the network parameter of every training step for evaluation and label guessing.

To have a fair comparison with MixMatch, we implemented an enhanced version of MixMatch by combing it with Augmentation Anchoring \cite{berthelot_remixmatch_2020}.
To report test accuracy, we take the checkpoint with the highest validation accuracy and report its test accuracy.
By default, our experiments have fixed hyperparameters $\tau_c=0.95$, $\tau_s=0.9$ and EMA decay to $0.999$.

\subsection{Datasets and Evaluations}

\textbf{CIFAR-10}: A dataset with 60K images of shape $32\times32$ evenly distributed across 10 classes.
The training set has 50K images, and the test set contains 10K images.
Our validation set size is 5000 for CIFAR-10.
The results are available in table \ref{table:CIFAR10-SVHN}.

\textbf{SVHN}: SVHN consists of 10 classes. Its training set has 73257 images, and the test set contains 26032 images.
Each image in SVHN is $32\times32$.
Our validation set size is 5000 for SVHN.
The results are available in table \ref{table:CIFAR10-SVHN}.

\textbf{CIFAR-100}: Similar to CIFAR-10, CIFAR-100 also has 50K training images and 10K test images but with 100 classes.
The image size is $32\times32$, the same as CIFAR-10.
Our validation set size is 5000 for CIFAR-100.
The results are available in table \ref{table:CIFAR100}.

\textbf{Mini-ImageNet}: Mini-ImageNet is first introduced in \cite{NIPS2016_MatchingNet} for few-shot learning.
The dataset contains 100 classes where each class has 600 images of size $84\times84$.
For SSL evaluation, our protocol follows that of \cite{Iscen2019_LabelPropagation}, in which 500 images are selected from each class to form the training set, and the leftover 100 images are used for testing.
Since \cite{Iscen2019_LabelPropagation} do not specify its validation set split, we use a total of 7200 training images (72 per class) as validation set; this is of the same validation set size as \cite{NIPS2016_MatchingNet}.

\textbf{DomainNet-Real} \cite{Peng2019_DomainNet}: DomainNet-Real has 345 categories with unbalanced numbers of images per class following a long tail distribution.
We use this dataset for transfer learning experiments in section \ref{section:transfer}.
For our evaluations, we resize the image to $84\times84$ and use 11-shot per class (a total of 3795) for the labeled training set.

\subsection{Baseline Methods}
We compare with the following baseline methods: FixMatch \cite{sohn_fixmatch_2020}, MixMatch \cite{berthelot_mixmatch_2019}, ReMixMatch \cite{berthelot_remixmatch_2020}, VAT \cite{Miyato2019_VAT}, MeanTeacher \cite{NIPS2017_MeanTeacher}, and Label Propagation \cite{Iscen2019_LabelPropagation}.

\input{tables/transfer1}

\subsection{Results}

% \subsubsection{Standard Benchmark}
For all datasets, our labeled and unlabeled set split is done by randomly sample the same number of images from all classes without replacement.
In general, our hyperparameter choices follows that of MixMatch \cite{berthelot_mixmatch_2019} and FixMatch \cite{sohn_fixmatch_2020}.

\textbf{CIFAR-100}: We set the loss weight to $\lambda_{\mathcal{U}}=150, \lambda_{\mathcal{P}}=150$.
As shown in table \ref{table:CIFAR100}, we find that SimPLE has significant improvement on CIFAR-100.
% Since \cite{berthelot_mixmatch_2019} only reported its CIFAR-100 test accuracy on a 26M parameter customized WRN 28 with 135 filters, we include experiments using a comparable network - WRN 28-8 with 23M parameters.
For better comparison with \cite{sohn_fixmatch_2020}, we include experiments using the same optimizer (SGD), hyperparameters, and backbone network (WRN 28-8 with 23M parameters).
With a larger backbone, our method still provides improvements over baseline methods.
SimPLE is better than FixMatch by 0.7\% and takes only 4.7 hours of training for convergence, while FixMatch takes about 8 hours to converge.
We consider convergence is achieved when the validation accuracy reaches 95\% of its highest value.

\textbf{CIFAR-10, SVHN}: For CIFAR-10, we set $\lambda_{\mathcal{U}}=75$ and $\lambda_{\mathcal{P}}=75$; we set $\lambda_{\mathcal{U}}=\lambda_{\mathcal{P}}=250$ for SVHN.
For both datasets, we use SGD with cosine learning rate decay \cite{loshchilov2016sgdr} with decay rate set to $\frac{7\pi}{16}$ following that of FixMatch \cite{sohn_fixmatch_2020}.

In table \ref{table:CIFAR10-SVHN}, we find that SimPLE is on par with ReMixMatch \cite{berthelot_remixmatch_2020} and FixMatch \cite{sohn_fixmatch_2020}.
ReMixMatch, FixMatch, and SimPLE are very close to the fully supervised baseline with less than 1\% difference in test accuracy.
SimPLE is less effective on these domains because the leftover samples are difficult ones whose pseudo labels are not similar to any of the high confidence pseudo labels.
In this case, no pseudo labels can pass the two thresholds in Pair Loss and contribute to the loss.
We observe that the percentage of pairs in a batch that passes both thresholds stabilizes early in the training progress (the percentage is 12\% for SVHN and 10\% for CIFAR-10).
Thus, Pair Loss does not bring much performance gain as it does in the more complicated datasets.

\textbf{Mini-ImageNet}: To examine the scalability of our method, we conduct experiments on Mini-ImageNet.
Mini-ImageNet is a more complex dataset because its categories and images are sampled directly from ImageNet.
Although the image size is scaled down to $84\times84$, it is still much more complicated than CIFAR-10, CIFAR-100, and SVHN.
Therefore, Mini-ImageNet is an excellent candidate to illustrate the scalability of SimPLE.

In addition to WRN 28-2 experiments on Mini-ImageNet, we also apply the SimPLE algorithm on ResNet-18 \cite{He2016_ResNet} for a fair comparison with prior works.
% All experiments on Mini-ImageNet use fixed augmentations instead of RandAugment.
The results are in table \ref{table:Mini-ImageNet}.
In general, our method outperforms all other methods by a large margin on Mini-ImageNet regardless of backbones.
Our method scales with the more challenging dataset.

\input{tables/transfer2}
\input{tables/ablation}

\subsubsection{SSL for Transfer Learning Task}
\label{section:transfer}

In real-world applications, a common scenario is where the target task is similar to existing datasets.
Transfer learning is helpful in this situation if the target domain has sufficient labeled data.
However, this is not guaranteed.
Therefore, SSL methods need to perform well when starting from a pre-trained model on a different dataset.
Another benefit of using a pre-trained model is having fast convergence, which is important for time-sensitive applications.

Since prior SSL methods often neglect this scenario, in this section, we evaluate our algorithm, MixMatch \cite{berthelot_mixmatch_2019} and supervised baseline in the transfer setting.
The supervised baseline only uses labeled training data and parameter EMA for evaluation.
All transfer experiments use fixed augmentations.

Our first experiment is the adaptation from DomainNet-Real to Mini-ImageNet; the result is in table \ref{table:transfer-Mini-ImageNet}.
We observe that the pre-trained models are on par with training from scratch but converge $5\sim 100$ times faster.
Under transfer setting, SimPLE is 7.57\% better than MixMatch and 9.9\% better than the supervised baseline.

The experiment in table \ref{table:transfer-DomainNet-Real} is for transferring from ImageNet-1K \cite{deng_2009_imagenet} to DomainNet-Real.
Since ImageNet-1K pre-trained ResNet-50 \cite{He2016_ResNet} is readily available in many machine learning libraries (e.g., PyTorch), we evaluate the performance and the convergence speed using ImageNet-1K pre-trained ResNet-50 to mimic real-world applications.

On DomainNet-Real, MixMatch is about 7\% lower than the supervised baseline, while SimPLE has 8\% higher accuracy than the baseline.
MixMatch Enhanced, despite having Augmentation Anchoring, does not outperform MixMatch.

It is clear that SimPLE perform well in pre-trained setting and surpasses MixMatch and supervised baselines by a large margin.
This behavior is consistent across datasets and network architectures.
MixMatch, on the other hand, does not improve performance in the pre-trained setting.

Compared to training from scratch, the pre-trained models do not always provide performance improvements since the pre-trained models might have domain bias that is not easy to overcome.
For example, in our DomainNet-Real to Mini-ImageNet experiment, the pre-trained test accuracy is slightly lower than training from scratch.
However, the convergence speed is significantly faster ($\sim$8 to 10 times) when starting from a pre-trained model.
% This property is helpful for time-sensitive applications.

\subsubsection{Ablation Study over CIFAR-100}
\label{section:ablation}

In this section, we conducted ablation studies on CIFAR-100 with WRN 28-2 to evaluate the effectiveness of different parts of our system.
The results are available in table \ref{table:ablation}.
We choose CIFAR-100 because it has a reasonable number of classes (reasonably complicated) and a small image size (fast enough for training).

We observe that Pair Loss significantly improves the performance.
With a more diverse augmentation policy or increasing the number of augmentations, the advantage of the Pair Loss is enhanced.
% Also, having a lower similarity threshold or confidence threshold does not have a significant performance impact.
Also, SimPLE is robust to threshold change.
One possible explanation for the robustness is that since a pair must pass both thresholds to contribute to the loss, changing one of them may not significantly affect the overall number of pairs that pass both thresholds.

%% file: tables/cifar100.tex
\begin{table}[!ht]
    \centering
    \begin{tabularx}{\columnwidth}{ c|c|c }
        \hline
        \multicolumn{3}{c}{CIFAR-100} \\
        \hline
        Method & 10000 labels & Backbone \\
        \hline\hline
        MixMatch$^*$ & 64.01\% & WRN 28-2 \\
        MixMatch Enhanced & 67.12\% & WRN 28-2 \\
        \textbf{SimPLE} & \textbf{70.82}\% & WRN 28-2 \\

        \hline
        MixMatch$^\dagger$ \cite{sohn_fixmatch_2020} & 71.69\% & WRN 28-8 \\
        ReMixMatch$^\dagger$ \cite{sohn_fixmatch_2020} & 76.97\% & WRN 28-8 \\
        FixMatch \cite{sohn_fixmatch_2020} & 77.40\% & WRN 28-8 \\
        % FixMatch + Pair Loss & 77.76\% & WRN 28-8 \\
        \textbf{SimPLE} & \textbf{78.11}\% & WRN 28-8 \\
        \hline
    \end{tabularx}
    \caption{
        CIFAR-100 Top-1 Test Accuracy.
        $^*$: using our implementation.
        $^\dagger$: reported in FixMatch \cite{sohn_fixmatch_2020}.
    }
    \label{table:CIFAR100}
\end{table}

%% file: tables/cifar10-svhn.tex
\begin{table*}[ht]
    \centering
    \begin{tabularx}{\textwidth}{ X|X X|X X }
        \hline
        & \multicolumn{2}{c|}{CIFAR-10} & \multicolumn{2}{c}{SVHN}\\
        \hline
        Method & 1000 labels & 4000 labels & 1000 labels & 4000 labels \\
        \hline\hline
        VAT$^\dagger$ \cite{berthelot_remixmatch_2020} & 81.36\% & 88.95\% & 94.02\% & 95.80\% \\
        MeanTeacher$^\dagger$ \cite{berthelot_remixmatch_2020} & 82.68\% & 89.64\% & 96.25\% & 96.61\% \\
        MixMatch \cite{berthelot_mixmatch_2019} & 92.25\% & 93.76\% & 96.73\% & 97.11\% \\
        ReMixMatch \cite{berthelot_remixmatch_2020}& 94.27\% & 94.86\% & 97.17\% & \textbf{97.58}\% \\
        FixMatch \cite{sohn_fixmatch_2020} & $-$ & \textbf{95.69\%} & \textbf{97.64\%} & $-$ \\
        \textbf{SimPLE} & \textbf{94.84\%} & 94.95\% & 97.54\% & 97.31\% \\
        \hline
        Fully Supervised$^{\dagger\ddagger}$ & \multicolumn{2}{c|}{95.75\%} & \multicolumn{2}{c}{97.3\%} \\
        \hline
    \end{tabularx}
    \caption{
        CIFAR-10 and SVHN Top-1 Test Accuracy. All experiments use WRN 28-2.
        $^\dagger$: The accuracy is reported in ReMixMatch \cite{berthelot_remixmatch_2020} and using its own implementation.
        $^\ddagger$: Fully supervised baseline using all the labels and simple augmentation (flip-and-crop).
    }
    \label{table:CIFAR10-SVHN}
\end{table*}

%% file: tables/mini-imagenet.tex
\begin{table*}[!ht]
    \centering
    \begin{tabularx}{\textwidth}{ X|X|X|X }
        \hline
        \multicolumn{4}{c}{Mini-ImageNet} \\
        \hline
        Method & 4000 labels & Backbone & K \\
        \hline\hline
        MixMatch$^*$ & 55.47\% & WRN 28-2 & 2 \\
        MixMatch Enhanced & 60.50\% & WRN 28-2 & 7 \\

        \textbf{SimPLE} & \textbf{66.55}\% & WRN 28-2 & 7 \\

        \hline
        MeanTeacher$^\dagger$ \cite{Iscen2019_LabelPropagation} & 27.49\% & Resnet-18 & -- \\
        Label Propagation \cite{Iscen2019_LabelPropagation} & 29.71\% & Resnet-18 & -- \\
        \textbf{SimPLE} & \textbf{45.51}\% & Resnet-18 & 7 \\
        \hline
    \end{tabularx}
    \caption{
        Mini-ImageNet Top-1 Test Accuracy.
        $^*$: using our implementation.
        $^\dagger$: The score is reported in \cite{Iscen2019_LabelPropagation} and using its own implementation.
    }
    \label{table:Mini-ImageNet}
\end{table*}

%% file: tables/transfer1.tex
\begin{table*}[!ht]
    \centering
    \begin{tabularx}{\textwidth}{ X|X|X }
        \hline
        \multicolumn{3}{c}{Transfer: DomainNet-Real to Mini-ImageNet} \\
        \hline
        Method & 4000 labels & Convergence step \\
        \hline\hline
        Supervised w/ EMA$^\S$ & 48.83\% & 4K \\
        \hline
        MixMatch$^*$ from scratch & 50.31\% & 150K \\
        MixMatch$^*$ & 53.39\% & 69K \\
        \hline
        MixMatch Enhanced$^*$ from scratch & 52.83\% & 734K \\
        MixMatch Enhanced$^*$ & 55.75\% & 7K\\
        \hline
        \textbf{SimPLE} from scratch & \textbf{59.92}\% & 338K \\
        \textbf{SimPLE} & \textbf{58.73}\% & 53K \\
        \hline
    \end{tabularx}
    \caption{
        DomainNet-Real pre-trained model transfer to Mini-ImageNet.
        All experiments use WRN 28-2.
        The model is converged when its validation accuracy reaches 95\% of its highest validation accuracy.
        $^\S$: using labeled training set only.
        $^*$: using our implementation.
    }
    \label{table:transfer-Mini-ImageNet}
\end{table*}

%% file: tables/transfer2.tex
\begin{table}[!ht]
    \centering
    \begin{tabularx}{\columnwidth}{ c|c|c }
        \hline
        \multicolumn{3}{c}{Transfer: ImageNet-1K to DomainNet-Real} \\
        \hline
        Method & 3795 labels & Convergence step \\
        \hline\hline
        Supervised w/ EMA$^\S$ & 42.91\% & 4K \\
        MixMatch$^*$ & 35.34\% & 5K \\
        MixMatch Enhanced$^*$ & 35.16\% & 5K \\
        \textbf{SimPLE} & \textbf{50.90}\% & 65K \\
        \hline
    \end{tabularx}
    \caption{
        ImageNet-1K pre-trained model transfer to DomainNet-Real.
        All experiments use ResNet-50.
        The model is converged when its validation accuracy reaches 95\% of its highest validation accuracy.
        $^\S$: using labeled training set only.
        $^*$: using our implementation.
    }
    \label{table:transfer-DomainNet-Real}
\end{table}

%% file: tables/ablation.tex
\begin{table*}[!ht]
    \centering
    \begin{tabularx}{\textwidth}{ X|X|c|c|c|c|c }
        \hline
        \multicolumn{7}{c}{Ablations: CIFAR-100} \\
        \hline
        Ablation & Augmentation Type & $\lambda_{\mathcal{P}}$ & $\tau_c$ & $\tau_s$ & K & 10000 labels \\
        \hline\hline
        \textbf{SimPLE} & RandAugment & 150 & 0.95 & 0.9 & 2 & 70.82\% \\
        \textbf{SimPLE} & RandAugment & 150 & 0.95 & 0.9 & \underline{7} &\textbf{73.04}\% \\
        w/o Pair Loss & RandAugment & \underline{0} & 0.95 & 0.9 & 2 & 69.07\% \\
        w/o Pair Loss & RandAugment & \underline{0} & 0.95 & 0.9 & \underline{7} & 69.94\% \\
        w/o RandAugment & \underline{fixed} & 150 & 0.95 & 0.9 & 2 & 67.91\% \\
        w/o RandAugment, w/o Pair Loss & \underline{fixed} & \underline{0} & 0.95 & 0.9 & 2 & 67.41\% \\
        \hline

        $\tau_c$ = 0.75 & RandAugment & 150 & \underline{0.75} & 0.9 & 2 & 71.96\% \\
        $\tau_s$ = 0.7 & RandAugment & 150 & 0.95 & \underline{0.7} & 2 & 70.85\%  \\
        $\tau_c$ = 0.75, $\tau_s$ = 0.7 & RandAugment & 150 & \underline{0.75} & \underline{0.7} & 2 & 71.48\% \\
        % $\tau_c$ = 0.55, $\tau_s$ = 0.5 & RandAugment & 150 & \underline{0.55} & \underline{0.5} & 2 & 71.25\% \\
        \hline

        $\lambda_{\mathcal{P}}$ = 50 & RandAugment & \underline{50} & 0.95 & 0.9 & 2 & 71.34\% \\
        $\lambda_{\mathcal{P}}$ = 250 & RandAugment & \underline{250} & 0.95 & 0.9 & 2 & 71.42\% \\
        \hline
    \end{tabularx}
    \caption{Ablation on CIFAR-100. All experiments use WRN 28-2}
    \label{table:ablation}
\end{table*}

%% file: sections/conclusion.tex
\section{Conclusion}
We proposed SimPLE, a semi-supervised learning algorithm. SimPLE improves on previous works \cite{berthelot_mixmatch_2019,berthelot_remixmatch_2020,sohn_fixmatch_2020} by considering a novel unsupervised objective, Pair Loss, which minimizes the statistical distance between high confidence pseudo labels with similarity above a certain threshold.
We have conducted extensive experiments over the standard datasets and demonstrated the effectiveness of the SimPLE algorithm.
Our method shows significant performance gains over previous state-of-the-art algorithms on CIFAR-100 and Mini-ImageNet \cite{NIPS2016_MatchingNet}, and is on par with the state-of-the-art methods on CIFAR-10 and SVHN.
Furthermore, SimPLE also outperforms the state-of-the-art methods in the transfer learning setting, where models are initialized by the weights pre-trained on ImageNet \cite{krizhevsky_imagenet_2012}, or DomainNet-Real \cite{Peng2019_DomainNet}.

%% file: sections/appendix.tex
\section{Implementation Detail}

\input{sections/appendix/hyperparameters}

\input{sections/appendix/augmentations}

\section{Further Analysis on Pair Loss}
\input{sections/appendix/threshold-analysis}
\input{sections/appendix/ablation}

%% file: sections/appendix/hyperparameters.tex
\subsection{Hyperparameters}

As mentioned in section \ref{section:experiments}, our hyperparameters are almost identical to that of MixMatch \cite{berthelot_mixmatch_2019} and FixMatch \cite{sohn_fixmatch_2020}.
We use the same network architecture and similar hyperparameters as FixMatch for CIFAR-10, SVHN, and CIFAR-100 (WRN 28-8).
We conducted ablation study on WRN 28-2 with hyperparameters similar to that of MixMatch for simplicity.
We also evaluated SimPLE on Mini-ImageNet with WRN 28-2 and ResNet 18.
We use the same $\alpha$ (beta distribution parameter for mix-up \cite{zhang_mixup_2018}) and $T$ (temperature for sharpening) across all experiments.
Notice that only MixMatch and MixMatch Enhanced use mix-up.

The full detail of our hyperparameters choices can be found in table \ref{table:hyperparameters1} and \ref{table:hyperparameters2}.
Our transfer experiment configurations are in table \ref{table:hyperparameters-transfer}.

\input{tables/appendix/hyperparameters1}

\subsection{Optimization}

For CIFAR-10, SVHN, and CIFAR-100 (WRN 28-8), we use SGD with Nesterov momentum set to $0.9$. We also use cosine learning rate decay \cite{loshchilov2016sgdr} with a decay rate of $\frac{7\pi}{16}$ following FixMatch.
For CIFAR-100 (WRN 28-2), Mini-ImageNet, and transfer experiments, we use AdamW \cite{loshchilov2018_AdamW} without learning rate scheduling follows that of MixMatch.
Details are available in table \ref{table:hyperparameters1}, \ref{table:hyperparameters2} and \ref{table:hyperparameters-transfer}.

\input{tables/appendix/hyperparameters2}
\input{tables/appendix/hyperparameters-transfer}

%% file: tables/appendix/hyperparameters1.tex
\begin{table}[!ht]
    \centering
    \begin{tabularx}{\columnwidth}{ c|c|c|c }
        \hline
        & CIFAR-10 & SVHN & CIFAR-100 \\
        \hline\hline
        $\tau_c$ & \multicolumn{3}{c}{0.95} \\
        $\tau_s$ & \multicolumn{3}{c}{0.9} \\

        \hline
        $\lambda_{\mathcal{U}}$ & 75 & 250 & 150 \\
        $\lambda_{\mathcal{P}}$ & 75 & 250 & 150 \\

        \hline
        $lr$ & \multicolumn{3}{c}{0.03} \\
        \hline
        $K$ & \multicolumn{2}{c|}{7} & 4 \\
        \hline
        $T$ & \multicolumn{3}{c}{0.5} \\
        $\alpha$ & \multicolumn{3}{c}{0.75} \\

        \hline
        weight decay & \multicolumn{2}{c|}{0.0005} & 0.001 \\
        \hline
        batch size & \multicolumn{3}{c}{64} \\
        \hline
        EMA decay & \multicolumn{3}{c}{0.999} \\

        \hline\hline
        backbone & \multicolumn{2}{c|}{WRN 28-2} & WRN 28-8 \\
        \hline
        optimizer & \multicolumn{3}{c}{SGD} \\
        Nesterov & \multicolumn{3}{c}{True} \\
        momentum & \multicolumn{3}{c}{0.9} \\
        \hline
        $lr$ scheduler & \multicolumn{3}{c}{cosine decay} \\
        $lr$ decay rate & \multicolumn{3}{c}{$7\pi \mathbin{/} 16$} \\
        \hline
    \end{tabularx}
    \caption{
        Hyperparameters for CIFAR-10, SVHN, and CIFAR-100 (with WRN 28-8).
    }
    \label{table:hyperparameters1}
\end{table}

%% file: tables/appendix/hyperparameters2.tex
\begin{table}[!ht]
    \centering
    \begin{tabularx}{\columnwidth}{ c|c|X|X }
        \hline
        & CIFAR-100 & \multicolumn{2}{c}{Mini-ImageNet} \\
        \hline\hline
        $\tau_c$ & \multicolumn{3}{c}{0.95} \\
        $\tau_s$ & \multicolumn{3}{c}{0.9} \\

        \hline
        $\lambda_{\mathcal{U}}$ & 150 & \multicolumn{2}{c}{300} \\
        $\lambda_{\mathcal{P}}$ & 150 & \multicolumn{2}{c}{300} \\

        \hline
        $lr$ & \multicolumn{3}{c}{0.002} \\
        \hline
        $K$ & 2 & \multicolumn{2}{c}{7} \\
        \hline
        $T$ & \multicolumn{3}{c}{0.5} \\
        $\alpha$ & \multicolumn{3}{c}{0.75} \\

        \hline
        weight decay & 0.04 & \multicolumn{2}{c}{0.02} \\
        \hline
        batch size & 64 & \multicolumn{2}{c}{16} \\
        \hline
        EMA decay & \multicolumn{3}{c}{0.999} \\

        \hline\hline
        backbone & WRN 28-2 & WRN 28-2 & ResNet 18 \\
        \hline
        optimizer & \multicolumn{3}{c}{AdamW} \\
        \hline
    \end{tabularx}
    \caption{
        Hyperparameters for CIFAR-100 (WRN 28-2) and Mini-ImageNet.
    }
    \label{table:hyperparameters2}
\end{table}

%% file: tables/appendix/hyperparameters-transfer.tex
\begin{table}[!ht]
    \centering
    \begin{tabularx}{\columnwidth}{ c|X|X }
        \hline
        & DN-R to M-IN & IN-1K to DN-R \\ 
        \hline\hline
        $\tau_c$ & \multicolumn{2}{c}{0.95} \\
        $\tau_s$ & \multicolumn{2}{c}{0.9} \\

        \hline
        $\lambda_{\mathcal{U}}$ & \multicolumn{2}{c}{300} \\
        $\lambda_{\mathcal{P}}$ & \multicolumn{2}{c}{300} \\

        \hline
        feature $lr$ & 0.0002 & 0.00002 \\
        \hline
        classifier $lr$ & \multicolumn{2}{c}{0.002} \\
        \hline
        $K$ & \multicolumn{2}{c}{2} \\
        $T$ & \multicolumn{2}{c}{0.5} \\
        $\alpha$ & \multicolumn{2}{c}{0.75} \\

        \hline
        weight decay & \multicolumn{2}{c}{0.02} \\
        \hline
        batch size & \multicolumn{2}{c}{16} \\
        \hline
        EMA decay & \multicolumn{2}{c}{0.999} \\

        \hline\hline
        backbone & WRN 28-2 & ResNet 50 \\
        \hline
        optimizer & \multicolumn{2}{c}{AdamW} \\
        \hline
    \end{tabularx}
    \caption{
        % Hyperparameters for DomainNet-Real to Mini-ImageNet (table \ref{table:transfer-Mini-ImageNet}) and ImageNet-1K to DomainNet-Real (table \ref{table:transfer-DomainNet-Real}) experiments.
        Hyperparameters for Transfer: DomainNet-Real to Mini-ImageNet (DN-R to M-IN) and Transfer: ImageNet-1K to DomainNet-Real (IN-1K to DN-R) experiments.
    }
    \label{table:hyperparameters-transfer}
\end{table}

%% file: sections/appendix/augmentations.tex
\subsection{Augmentations}

\input{tables/appendix/augmentations}

Our augmentations are implemented on GPU with Kornia \cite{eriba2019kornia}.
In table \ref{table:augmentations}, we list the transformations used by the fixed augmentations of table \ref{table:transfer-Mini-ImageNet} and \ref{table:transfer-DomainNet-Real}.
For RandAugment \cite{Cubuk2020_RandAugment}, we follows the exact same settings as FixMatch \cite{sohn_fixmatch_2020}.
Note that we only reported the changed augmentation parameters while the omitted values are the same as the default parameters in Kornia \cite{eriba2019kornia}.

%% file: tables/appendix/augmentations.tex
\begin{table*}[!ht]
    \centering
    \begin{tabularx}{\textwidth}{ c|X|X }
        \hline
        Transformation & Description & Parameter \\
        \hline\hline
        Random Horizontal Flip & Horizontally flip an image randomly with a given probability $p$ & $p=0.5$ \\
        \hline
        Random Resized Crop & Random crop on given size and resizing the cropped patch to another & scale $=(0.8, 1)$, ratio $=(1, 1)$ \\
        \hline
        Random 2D GaussianBlur & Creates an Gaussian filter for image blurring. The blurring is randomly applied with probability $p$ & $p=0.5$, kernel size $=(3, 3)$, sigma $=(1.5, 1.5)$\\
        \hline
        Color Jitter & Randomly change the brightness, contrast, saturation, and hue of given images & contrast $=(0.75, 1.5)$\\
        \hline
        Random Erasing & Erases a randomly selected rectangle for each image in the batch, putting the value to zero & $p=0.1$\\
        \hline
        Random Affine & Random affine transformation of the image keeping center invariant & degrees $=(-25, 25)$, translate $=(0.2, 0.2)$, scale $=(0.8, 1.2)$, shear $=(-8, 8)$ \\
        \hline
    \end{tabularx}
    \caption{
        Augmentation details. Applied in order. Descriptions are from \cite{eriba2019kornia}.
    }
    \label{table:augmentations}
\end{table*}

%% file: sections/appendix/threshold-analysis.tex
\subsection{Analysis on Confidence Threshold}

\newcommand{\lpnorm}[2]{\left\lVert#2\right\rVert_{#1}}
\newcommand{\geodist}[2]{d_{S_n}\left({#1}, {#2}\right)}

\begin{theorem}
$\forall p, q \in \Delta^N$, if $\thresholdFunc{\max\left(p\right)}{\tau_c} \cdot \thresholdFunc{\simFunc{p}{q}}{\tau_s} > 0$, then $\max \left( q \right) > \cos(\cos^{-1}(\sqrt{\tau_c})+\cos^{-1}(\tau_s))^2$.
\end{theorem}

Since $\thresholdFunc{\max\left(p \right)}{\tau_c} \cdot \thresholdFunc{\simFunc{p}{q}}{\tau_s} > 0$, we have:
\begin{equation*}
    \begin{cases}
    \max \left( p \right) > \tau_c \\
    \simFunc{p}{q} > \tau_s
    \end{cases}
\end{equation*}

Denote $j = \arg \max_i p_i$, i.e., the confidence of $p$ is attained at the $j$-th coordinate, $p_j = \max (p)$.

Denote $e_j \in \Delta^n$ as the elementary vector with the $j$-th element to be $1$ and all other elements to be $0$.

In the square root probability space, we have:
\begin{equation*}
    \begin{cases}
    \sqrt{e_j}^\top \sqrt{p} = \max \left( \sqrt{p} \right) &> \sqrt{\tau_c}\\
    \sqrt{p}^\top \sqrt{q} > \tau_s
    \end{cases}
\end{equation*}

Notice, because $\lpnorm{1}{p} = \lpnorm{1}{q} = \lpnorm{1}{e_j} = 1$, we have $\lpnorm{2}{\sqrt{p}} = \lpnorm{2}{\sqrt{q}} = \lpnorm{2}{\sqrt{e_j}} = 1$. Therefore, $\sqrt{p}$, $\sqrt{q}$, and $\sqrt{e_j}$ are on the unit $n$-sphere $S_n$. Denote the geodesic distance between any two points $x, y \in S_n$ as $\geodist{x}{y} = \cos^{-1}(\frac{x^\top y}{\lpnorm{2}{x} \cdot \lpnorm{2}{y}}) = \cos^{-1}(x^\top y)$ .

% Therefore, $\frac{\sqrt{e_j}^\top \sqrt{p}}{\lpnorm{2}{\sqrt{e_j}} \lpnorm{2}{\sqrt{p}}} > \sqrt{\tau_c}$ and $\frac{\sqrt{p}^\top \sqrt{q}}{ \lpnorm{2}{\sqrt{p}} \lpnorm{2}{\sqrt{q}} } > \tau_s$.
\begin{equation*}
    \begin{cases}
    \geodist{\sqrt{p}}{\sqrt{e_j}} > \cos^{-1}(\sqrt{\tau_c})\\
    \geodist{\sqrt{p}}{\sqrt{q}} > \cos^{-1}(\tau_s)
    \end{cases}
\end{equation*}

As the geodesic distance preserves triangular inequality:
\begin{align*}
    \geodist{\sqrt{q}}{\sqrt{e_j}} &\geq \geodist{\sqrt{q}}{\sqrt{p}} + \geodist{\sqrt{p}}{\sqrt{e_j}}\\
    &> \cos^{-1}(\sqrt{\tau_c}) + \cos^{-1}(\tau_s) \\
    \sqrt{q_j} = \sqrt{q}^\top \sqrt{e_j} &> \cos (\cos^{-1}(\sqrt{\tau_c}) + \cos^{-1}(\tau_s)) \\
    \max(q) \geq q_j &> \cos (\cos^{-1}(\sqrt{\tau_c}) + \cos^{-1}(\tau_s))^2
\end{align*}

%% file: sections/appendix/ablation.tex
\subsection{More on Pair Loss}

In this section, we provide additional information to two existing ablation studies in table \ref{table:ablation} on CIFAR-100,
to demonstrate the effectiveness of Pair Loss in encouraging more unlabeled samples to have accurate and high confidence predictions.
Specifically, we compare the performance of the SimPLE algorithm with and without the Pair Loss enabled in the following measurements:
1) the percentage of unlabeled samples with high confidence pseudo labels;
2) the percentage of unlabeled sample pairs that pass both confidence and similarity thresholds;
3) the percentage of false-positive unlabeled sample pairs that pass both confidence and similarity thresholds but are in different categories.

\begin{figure}[H]
    \centering
    \includegraphics[width=\columnwidth]{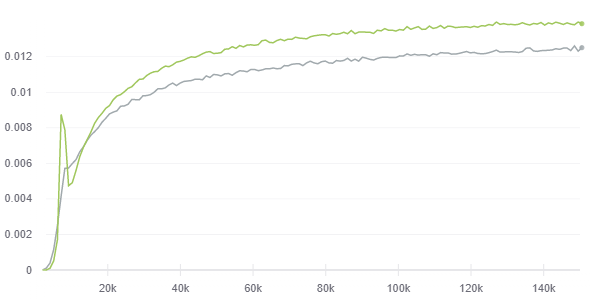}
    \caption{Ratio of pairs pass both confidence and similarity thresholds. The green line is SimPLE and the grey line is SimPLE without Pair Loss}
    \label{fig:thresholded-ratio}
\end{figure}

From figure \ref{fig:thresholded-ratio}, the ratio of pairs that pass both the confidence threshold and similarity threshold is increased by 16.67\%, with a consistently nearly 0\% false positive rate, which indicates that Pair Loss encourages the model to make more consistent and similar predictions for unlabeled samples from the same class.

\begin{figure}[H]
    \centering
    \includegraphics[width=\columnwidth]{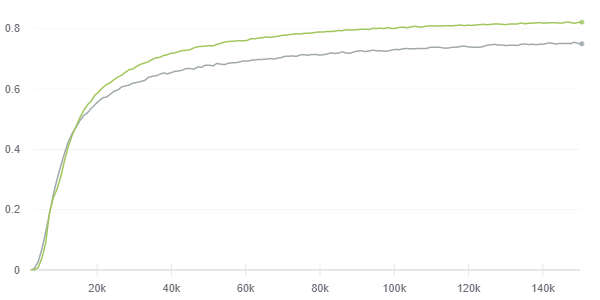}
    \caption{Ratio of high confidence prediction. The green line is SimPLE and the grey line is SimPLE without Pair Loss}
    \label{fig:high-conf-ratio}
\end{figure}

As shown in figure \ref{fig:high-conf-ratio}, with Pair Loss, the percentage of unlabeled sample with high confidence labels is increased by 7.5\%, and the prediction accuracy is increased by 2\% as shown in table \ref{table:ablation}.
These two results indicate that Pair Loss encourages the model to make high confidence and accurate predictions on more unlabeled samples, which follows our expectation that Pair Loss aligns samples with lower confidence pseudo labels to their similar high confidence counterparts during the training and improves the prediction accuracy.